\documentclass[conference]{IEEEtran}
\IEEEoverridecommandlockouts
\usepackage{cite}
\usepackage{amsmath,amssymb,amsfonts}
\usepackage{algorithmic}
\usepackage{graphicx}
\usepackage{textcomp}
\usepackage{xcolor}

\usepackage{booktabs}
\usepackage[colorlinks=true,allcolors=black]{hyperref}

\def\BibTeX{{\rm B\kern-.05em{\sc i\kern-.025em b}\kern-.08em
    T\kern-.1667em\lower.7ex\hbox{E}\kern-.125emX}}
\begin{document}

\title{Large Model based Sequential Keyframe Extraction for Video Summarization}
%

\author{
	\IEEEauthorblockN{Kailong Tan$^{\dagger*}$, Yuxiang Zhou$^{\heartsuit*}$, Qianchen Xia$^\ddagger$\thanks{*equal contributions, $\ddagger$corresponding author.}, Rui Liu$^{\diamondsuit}$, Yong Chen$^\heartsuit$}
	\IEEEauthorblockA{$^\dagger$School of Computer Science, China University of Geosciences, Beijing, China}
	\IEEEauthorblockA{$^\heartsuit$School of Computer Science, Beijing University of Posts and Telecommunications, Beijing, China}
	\IEEEauthorblockA{$^\ddagger$School of Mechanical Engineering, Tsinghua University, Beijing, China}
	\IEEEauthorblockA{$^\diamondsuit$School of Computer Science and Engineering, Beihang University, Beijing, China}
	\IEEEauthorblockA{kailong.tan@email.cugb.edu.cn, yuxiang.zhou@bupt.edu.cn, qianchenxia@tsinghua.edu.cn, lr@buaa.edu.cn}
}

\maketitle

\begin{abstract}
Keyframe extraction aims to sum up a video's semantics with the minimum number of its frames. 
This paper puts forward a \underline{L}arge \underline{M}odel based \underline{S}equential \underline{K}eyframe \underline{E}xtraction for video summarization, dubbed \textbf{LMSKE}, which contains three stages as below. 
First, we use the large model ``TransNetV2\footnote{\url{https://github.com/soCzech/TransNetV2}}'' to cut the video into consecutive shots, and employ the large model ``CLIP\footnote{\url{https://github.com/openai/CLIP}}'' to generate each frame's visual feature within each shot; 
Second, we develop an adaptive clustering algorithm to yield candidate keyframes for each shot, with each candidate keyframe locating nearest to a cluster center;
Third, we further reduce the above candidate keyframes via redundancy elimination within each shot, and finally concatenate them in accordance with the sequence of shots as the final sequential keyframes.
To evaluate LMSKE, we curate a benchmark dataset and conduct rich experiments, whose results exhibit that LMSKE performs much better than quite a few SOTA competitors with average F1 of 0.5311, average fidelity of 0.8141, and average compression ratio of 0.9922.
\end{abstract}


\section{Introduction}
With the popularity of Youtube and Tiktok platforms, video has become one of the mainstream media in our everyday life, communication and learning. 
Video keyframe extraction aims to draw as few temporal frames as possible from a given video to summarize its visual semantics, 
which is a key foundational technology for video storage, retrieval, and analysis~\cite{MCMSH-Yanbin-ACMMM-2022}. 
With the support of large models~\cite{InstructGPT-Long-NeurIPS-2022,openaiGPT4TechnicalReport2023}, this technology is also receiving increasing attention.

\begin{figure*}[ht]
	\centering
	\includegraphics[width=0.95\textwidth]{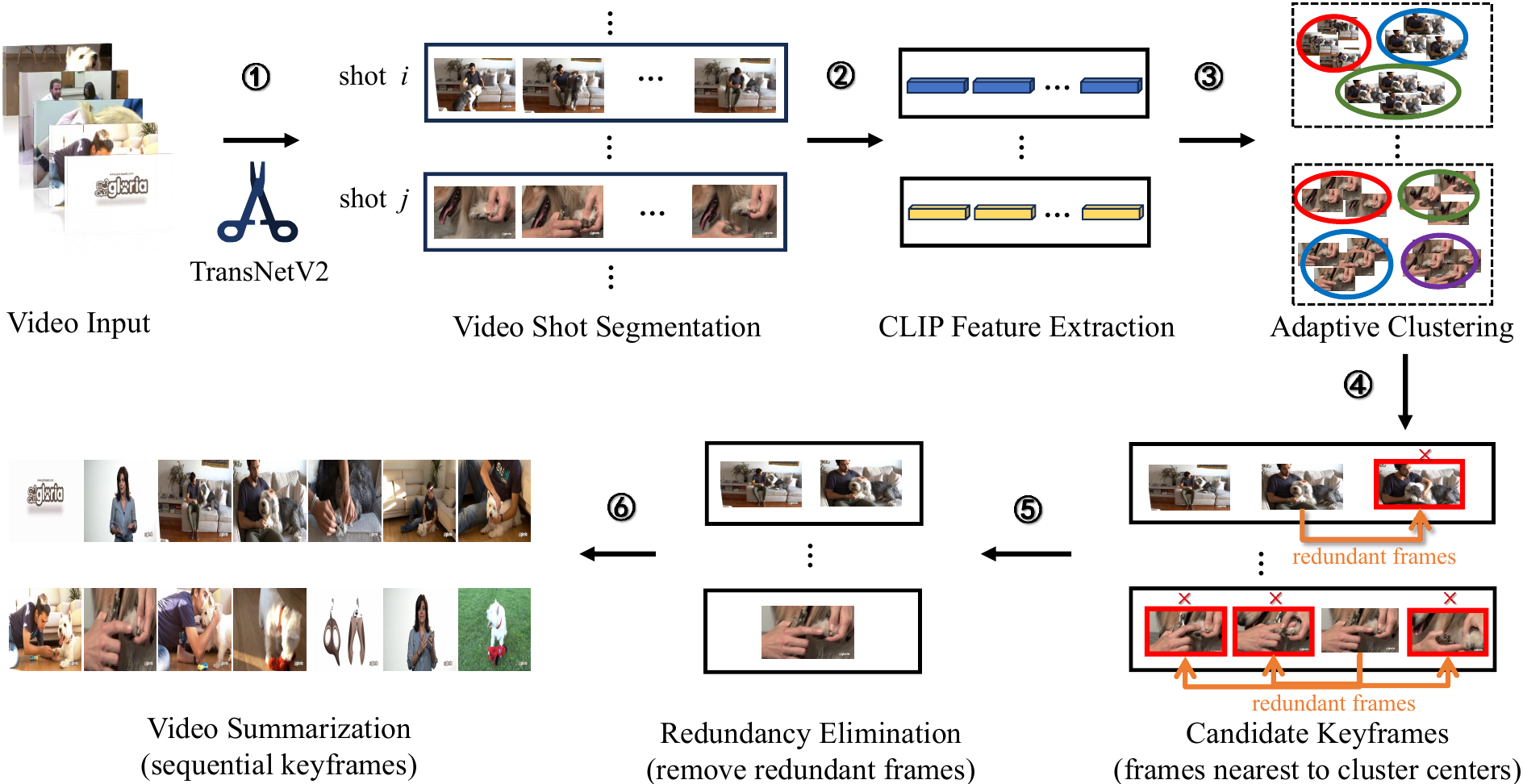}
	\caption{Our LMSKE framework: shot segmentation, adaptive clustering, and redundancy elimination.}
	\label{fig:architecture}
\end{figure*}

Currently, the existing keyframe extraction methods could be categorized into four classes: (1) uniform sampling based, (2) clustering based, (3) comparison based, (4) shot based approaches.

For the first class, keyframes are sampled uniformly from a video at a pre-defined interval, which is efficient but low effective because it's uncertain whether the extracted keyframes are redundant or sufficient to express videos' semantics.

For the second class, video frames are organized into clusters, and from each the most representive frames are selected as keyframes. 
For example, VSUMM~\cite{VSUMM-Sandra-PatternRecognitLett-2011}, SGC~\cite{clustering-Mingjun-CISAT-2021}, GMC~\cite{GraphModularityClustering-Hana-ICASSP-2017} used k-means, minimum spanning tree, and graph modularity based clustering algorithms for keyframe extraction, respectively.
The extracted keyframes from these methods can well reflect the contents of videos, but they ignored the keyframes' temporal sequences. 
Besides, the optimal number of clusters is always challenging to be decided within various videos.

For the third class, such as VSUKFE~\cite{FrameDifference-YiyinDing-Symposia-2021} and DiffHist~\cite{AbsoluteDifference-Rodriguez-ICALIP-2018}, they detect sudden changes in distances between consecutive frames, and recognize the keyframes only when the difference between frames exceeds a certain threshold. 
Such approaches keep their temporal relationships but the threshold is hard to set under different conditions.

For the fourth class, they extract keyframes from each shot and concatenate them~\cite{ShotSegment-Nandini-King-2022,ShotSegment-Kumar-EngineeringSystems-2020}. These methods can also maintain keyframes' sequences, but drawing only one frame from each shot is insufficient to fully describe videos' visual contents; 
in addition, using traditional features for boundary detection might be inaccurate for shot segmentations.

Overall, the extracted keyframes of the above methods fail to achieve a good balance between precision, recall, and temporal order against the benchmark keyframes. 
To handle such problem, we present a large model based sequential keyframe extraction, dubbed {LMSKE}, to extract minimal keyframes to sum up a given video with their sequences maintained. 
First, the large model TransNetV2~\cite{TransNetv2-Tom-CoRR-2020} was utilized to conduct shot segmentations, and the large model CLIP~\cite{CLIP-Alec-ICML-2021} was employed to extract semantic features for each frame within each shot. 
Second, an adaptive clustering method is devised to automatically determine the optimal clusters, based on which we performed candidate keyframe selection and redundancy elimination shot by shot. 
Finally, a keyframe set was obtained by concatenating keyframes of all shots in chronological order.

The contributions of this work can be listed as follows:
\begin{itemize}
	\item We proposed a large model based sequential keyframe extraction method, which can identify sequential keyframes in 3 stages: shot segmentation, adaptive clustering, and redundancy elimination (Fig.~\ref{fig:architecture}).
	
	\item We curated a benchmark dataset named TVSum20, and opened it to the public with the aim to promote the development of keyframe extraction approaches.
	
	\item We conducted rich experiments on TVSum20 to demonstrate that the proposed approach can better summarize videos with fewer frames than SOTA competitors.
\end{itemize}

\section{Method}\label{sec:method}

Fig.~\ref{fig:architecture} illustrates our LMSKE solution for video summarization, described step-by-step as below.

\subsection{Shot Segmentation and Feature Exrtaction}

Given a video $\mathcal{V} = \{\mathbf{x}_i\}_{i=1}^{l}$, $\mathbf{x}_i$ denotes its $i$-th frame, and $l$ is the total number of all its frames.  
Feeding $\mathcal{V}$ into the large model TransNetV2~\cite{TransNetv2-Tom-CoRR-2020}, video boundaries $\mathcal{B} = \{\left(s_i,e_i\right)\}_{i=1}^{m}$ are derived and then utilized to partition the video into a set of temporal shots $ \{\mathcal{S}_1,\mathcal{S}_2,\cdots,\mathcal{S}_m\}$, where $m$ represents the total number of video shots, $s_i$ and $e_i$ denote the start and end index of the $i$-th shot, respectively; thereafter, $\mathcal{S}_i = \{\mathbf{x}_j\}_{j=s_i}^{e_i}$ marks the $i$-th shot of the video $\mathcal{V}$.

In what follows, for each shot $\mathcal{S}_i$, we feed it into the large model CLIP~\cite{CLIP-Alec-ICML-2021} and obtain shot features $\mathbf{F}_i = [\mathbf{f}_{s_i},\cdots, \mathbf{f}_{e_i}]$, where $\mathbf{f}_j \in \mathbb{R}^{768}$ marks its $j$-th frame's semantic vector. 
Here, it is worth mentioning that, compared with the conventional features such as SIFT ~\cite{KunweiSong-SIFT-FAIML-2022}and HOG ~\cite{YanhuiXu-HOG-IAEAC-2022}, the deep features of CLIP encompass richer semantics, leading to better clustering results. 
$\mathbf{F}_i$ are then used as inputs to the following adaptive clustering algorithm (in section~\ref{sec:adaptive-clustering}) that partitions the shot frames into groups for preliminary keyframes selection. 

\subsection{Adaptive Clustering}\label{sec:adaptive-clustering}

\begin{figure}[t]
	\centering
	\includegraphics[width=8.5cm]{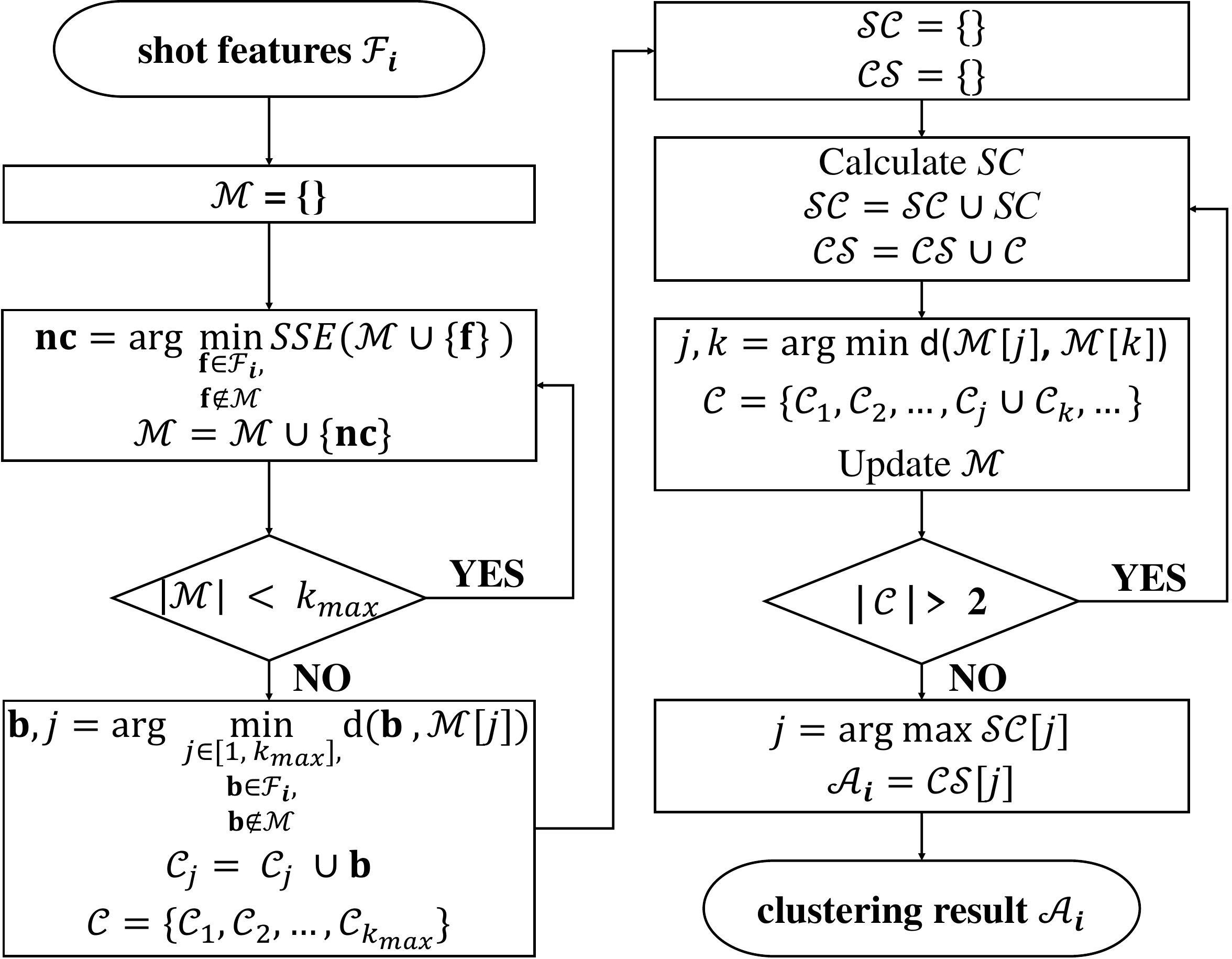}
	\caption{The flow chart of the adaptive clustering algorithm.}
	\label{fig:Adaptive clustering}
\end{figure}

We devise a k-means based clustering method, which can automatically determine the optimal number of clusters, to cluster each shot features $\mathbf{F}_i$ into groups for deriving candidate keyframes $\mathcal{A}_i$.
Its algorithm's flowchart is illustrated in Fig.~\ref{fig:Adaptive clustering}.

Concretely, the initial cluster centers are obtained by minimizing $SSE$ (Sum of Squared Errors)~\cite{Nainggolan-sse-journal-2019}, where $SSE$ denotes the sum of squared distances between each data point and its cluster center. 
Clearly, a smaller $SSE$ value suggests a more compact clustering results. $SSE$ can be formulated as:
\begin{equation}\label{eq:SSE}
	SSE(\mathcal{M}) = \sum_{k=1}^{n} \sum_{j=1}^{|\mathcal{M}|} (\mathbf{x}_{kj} - \mathcal{C}_{j})^2,
\end{equation}
where $n$ represents the total number of frames in shot $\mathcal{S}_i$, 
$\mathcal{M}$ denotes the cluster center set (thus $|\mathcal{M}|$ marks the number of cluster centers),
$\mathbf{x}_{kj}$ denotes the feature vector of the $k$-th frame belonging to the $j$-th cluster, and $\mathcal{C}_j$ represents the center of the $j$-th cluster.

According to~\cite{Kmax-Shi-EURASIP-2021}, we set the initial number of cluster centers as $k_{max} =|\mathcal{M}|= \sqrt{n}$. 
As Fig.~\ref{fig:Adaptive clustering} shows, at the begining, $\mathcal{M}$ is initialized empty. 
Our goal is to find $k_{max}$ cluster centers via $k_{max}$ iterations. 
In each iteration, we assume each data point except $\mathcal{M}$ as a potential new cluster center, and calculate its corresponding $SSE$. 
Subsequently, we select the data point with the minimum $SSE$ as a new cluster center $\mathbf{nc}$ and add it to $\mathcal{M}$. 
Repeat the above process until the number of cluster centers reaches $k_{max}$. 
Then, the shot $S_i$ was partitioned into $k_{max}$ clusters $\{\mathcal{C}_{1},\mathcal{C}_{2},\cdots,\mathcal{C}_{k_{max}}\}$. 

Next, the optimal clustering result is achieved by maximizing the $SC$ (silhouette coefficient)~\cite{GuoJinHeng-SC-ICCEAI-2022}, defined as:
\begin{equation}
	SC = \frac{1}{n} \sum_{i=1}^{n} S(i), \label{eq:SC}
\end{equation}
where
\begin{equation}
S(i) = \frac{b(i) - a(i)}{\max\{a(i), b(i)\}}, \label{eq:Si}
\end{equation}
and $a(i)$ denotes the average distance of the data point $i$ to the other data points within the same cluster, $b(i)$ marks the minimum distance of the data point $i$ to the other cluster centers.

To be specific, we first merge two closest clusters into one cluster and the mean of all data points within these two clusters is employed as the new cluster center. 
At the same time, the $SC$ is computed and the corresponding cluster centers are recorded. 
Repeat this process until only one cluster remains. 
Then the cluster centers with the highest $SC$ are chosen as the final clustering. 
The frames closest to the cluster centers are assembled to form a collection of candidate keyframe set $\mathcal{A}_i$.

\subsection{Redundancy Elimination}
In practise, $\mathcal{A}_i$ may contain redundant frames or uninformative (e.g., solid-color) frames.
 
To start with, we employ the color histograms (8$\times$8$\times$8 in HSV channels~\cite{ImaKurniastuti-HSV-IAEAC-2021}) of candidate keyframes to remove solid-color or uninformative frames. 
More specifically, frames with histogram non-zero bin counts below 10 are identified as uninformative or solid-color frames. 

Next, by computing the similarity between any two candidate keyframes based on color histograms, we build a similarity matrix $\mathbf{SIM}$. We traverse $\mathbf{SIM}$ and find the two frames with the highest similarity: 
\begin{equation}
	i,j = \arg\max_{i,j}\text{Triu}(\mathbf{SIM}[i,j]),
\end{equation}
where Triu($\cdot$) is employed to extract the upper triangular part of matrix $\mathbf{SIM}$. 
Then we eliminate the $j$-th frame, which is viewed as redundant, from $\mathcal{A}_i$ and update $\mathbf{SIM}$ by eliminating the $j$-th column. 
Iterate this process until the maximum similarity falls below the specified threshold $0.8$. 

Ultimately, after redundancy elimination of the candidate keyframe set $\mathcal{A}_i$, a compact keyframe set $\mathcal{K}_i$ is obtained for shot $\mathcal{S}_i$. 
Arrange the keyframe sets of all shots in chronological order to obtain the keyframe set $\mathcal{K}$ of the entire video $\mathcal{V}$:
\begin{equation}
	\mathcal{K} = \oplus_{i=1}^{m} \mathcal{K}_i, \label{eq:final keyframe set}
\end{equation}
where $\oplus$ signifies concatenation in sequence.

\section{EXPERIMENT}

In this section, we will validate the effectiveness of our proposed LMSKE approach both quantitatively and qualitatively.

\subsection{Dataset}
The existing evaluations of keyframe extraction approaches are based on experts' interpretation of extracted keyframes, which is not only expensive but also subjective;
thus, we curated a benchmark dataset based on TVSum\footnote{\url{http://people.csail.mit.edu/yalesong/tvsum/}}~\cite{TVSum-Yale-CVPR-2015}.
 
TVSum holds 50 videos collected from YouTube, spanning across 10 distinct categories. 
For every 2 seconds of one video, an important score is assigned by 20 experts. 
We analyze that the average important scores could serve as potential scores for identifying keyframes. 
Specifically, we distinguish all 2 seconds' segments with local maximum scores and select its central frame as a candidate keyframe. In what follows, we manually remove the redundant frames and low-information frames shot by shot. 
Finally, we build a benchmark dataset, dubbed \textbf{TVSum20}, dedicated to evaluating the performance of keyframe extraction methods.

TVSum20 contains 20 videos, i.e., 10 different categories with each one covering 2 videos, where each video owns sequental keyframes. 
It's open to the public on github \url{https://github.com/ttharden/Keyframe-extraction}.

\begin{figure*}[t]
	\centering
	\includegraphics[width=0.90\textwidth]{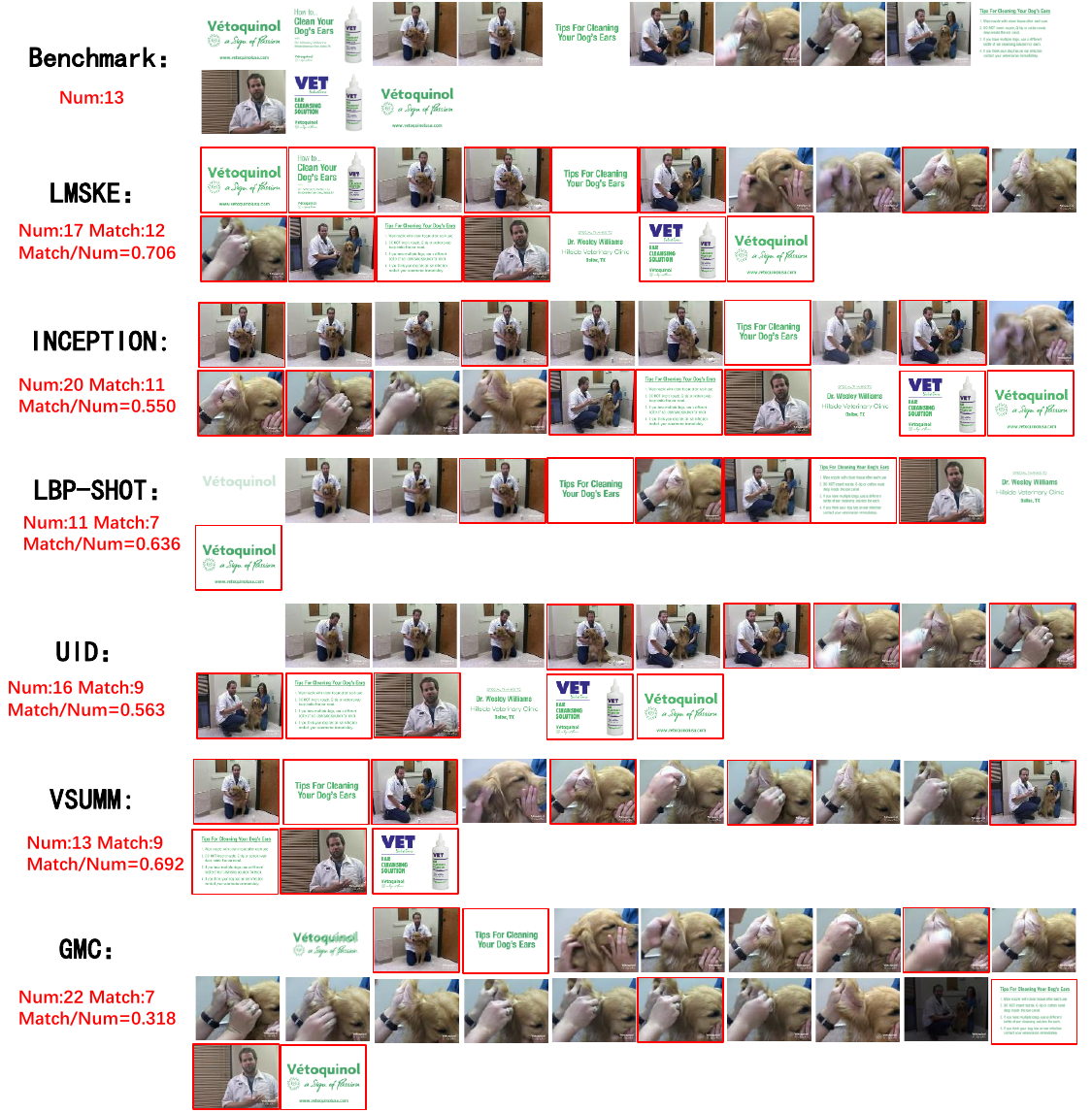}
	\caption{Qualitative comparisons between the benchmark and the representative methods (such as LMSKE, INCEPTION, LBP-SHOT, UID, VSUMM, and GMC). Note that Uniform and K-Means (appeared in Table~\ref{tab:experiment_result}) are not illustrated here because the numbers of their selected keyframes are relatively large and obviously they perform worse than other competitors.}
	\label{fig:case-study}
\end{figure*}

\subsection{Metrics and Competitors}
We evaluated the keyframe extraction approaches using three popular metrics: F1~\cite{SHOT-Nandini-JournalofKingSaudUniversity-2022,Review-Milan-CS-2018}, 
Fidelity~\cite{SHOT-Nandini-JournalofKingSaudUniversity-2022}, 
and CR (compression ratio)~\cite{GraphModularityClustering-Hana-ICASSP-2017}.

F1 is an indicator comprehensively considering the precision and recall of the extracted keyframes compared with the benchmarks, and if a method could yield a larger F1 score, it will extract higher-quailty keyframes.

Higher fidelity values indicate that the extracted keyframe set provides a better global description of the visual content of the given video. 

CR is used to study the compactness of the keyframe set $\mathcal{K}$ and it depends on the number of selected keyframes given a video $\mathcal{V}$. If a method could yield a larger CR value, it will extract fewer keyframes to summarize the video.

For each video, we can compute one F1, Fidelity and CR value, and the average values over all videos are recorded as the final results.

We compared our method with quite a few representative keyframe extraction methods, i.e.,
Uniform (30 frames)~\cite{Hao-DeepKeyFrame-ACM-2023},
DiffHist~\cite{AbsoluteDifference-Rodriguez-ICALIP-2018},
VSUMM~\cite{VSUMM-Sandra-PatternRecognitLett-2011},
K-Means~\cite{Muhammad-Kmeans-IJIPC-2020},
GMC~\cite{Dateset1},
UID~\cite{VideoSum-Garcia-CRAS-2023},
INCEPTION~\cite{VideoSum-Garcia-CRAS-2023},
and LBP-Shot~\cite{SHOT-Nandini-JournalofKingSaudUniversity-2022}.

\subsection{Results}

\begin{table}[htbp]
        \caption{Average F1, Fidelity, CR values of various methods.}
	\begin{center}
        \begin{tabular}{|c|c|c|c|}
        \toprule
		\multicolumn{1}{|c|}{Methods}   & {F1} & {Fidelity} & {CR} \\ \midrule
		{Uniform}         & 0.2061     & 0.7264            & 0.9662      \\  
		{VSUMM}     & 0.4894     & 0.7919            & 0.9909      \\ 
		{K-Means}    & 0.5039     & 0.7975            & 0.9895      \\ 
		{GMC}    & 0.4833     & 0.7854            & 0.9883      \\ 
		{DiffHist}    & 0.3380      & 0.7696            & 0.9835      \\
		{UID}       & 0.4615     & 0.7872            & 0.9902      \\ 
		{INCEPTION} & 0.5168     & 0.7906            & 0.9908      \\ 
		{LBP-SHOT}      & 0.5050      & 0.7967            & 0.9910      \\ \midrule
		{LMSKE}              & \textbf{0.5311}     & \textbf{0.8141}            & \textbf{0.9922}      \\
		\bottomrule
            \end{tabular}
        \end{center}
	
	\label{tab:experiment_result}
\end{table}

We have conducted extensive experiments on TVSum20 and collected their results in Table~\ref{tab:experiment_result}. 
As can be seen, no matter it's F1, Fidelity or CR, LMSKE yields the highest values;
specifically, compared to the most competitive method INCEPTION, LMSKE still outperforms it by an improvement of 2.77\%, 2.97\%, and 0.14\% with respect to F1, Fidelity, and CR respectively, which suggests that our LMSKE could better summarize the video contents with fewer video frames than other methods.

\subsection{Case Study}
Fig.~\ref{fig:case-study} shows a video's keyframes extracted by the benchmark and 6 well-performed methods, where a frame in red border indicates a match with the benchmark. 
Apparently, LMSKE extract 12 matches, the most, which suggests the highest recall; 
besides, by comparing $Match/Num$, LMSKE yields the largest value, which suggests the highest precison.
In addition, we can clearly observe that the keyframe sequence of LMSKE well matches that of the Benchmark. 

\section{Conclusion}
\label{sec:conclusion}

This paper proposes a three-stage sequential keyframe extraction approach for video summarization, and differs from the current approaches in 3 aspects: 
(1) leverage large models to cut video into high-quality shots and embed each frame with a semantic vector for better clustering;
(2) design an adaptive clustering algorithm to divide each video shot into several clusters for initial keyframe generation;
(3) further eliminate redundant keyframes shot by shot.
It is worth noting that, we have built a standard dataset and made it publicly available for the evaluation of keyframe extraction methods. 
Experiments validate that our LMSKE could capture more semantics with fewer video frames than other competitors.

\section{Acknowledgement}
\label{sec:Acknowledgement}

The authors would like to thank the reviewers and the editor
for their constructive comments. This work was supported in
part by National Natural Science Foundation of China (Grant No.~62372054) and National Key Research and Development
Program of China (Grant No.~2022YFC3302200).

\bibliographystyle{IEEEbib}
\bibliography{refs}

\end{document}